%% file: main.tex
\documentclass[10pt,twocolumn,letterpaper]{article}

\usepackage{cvpr}              %

\input{packages}
\input{macros}

\def\paperID{846} %
\def\confName{CVPR}
\def\confYear{2024}

\title{Lazy Diffusion Transformer for Interactive Image Editing}

\author{
 Yotam Nitzan$^{1,2}$ \hspace{1em} \and
 Zongze Wu$^1$ \hspace{1em} \and
 Richard Zhang$^1$ \hspace{1em} \and
 Eli Shechtman$^1$ \hspace{1em} \and
 Daniel Cohen-Or$^2$ \hspace{1em} \and
 Taesung Park$^{1}$ \hspace{1em} \and 
 Michaël Gharbi$^{1}$ \hspace{1em} \and \\
 $^{1}$ Adobe Research \qquad $^2$ Tel-Aviv University
 \\ \url{https://lazydiffusion.github.io}
}

\begin{document}

\twocolumn[{%
    \maketitle
    \begin{center}
        \centering
        \includegraphics[width=\linewidth]{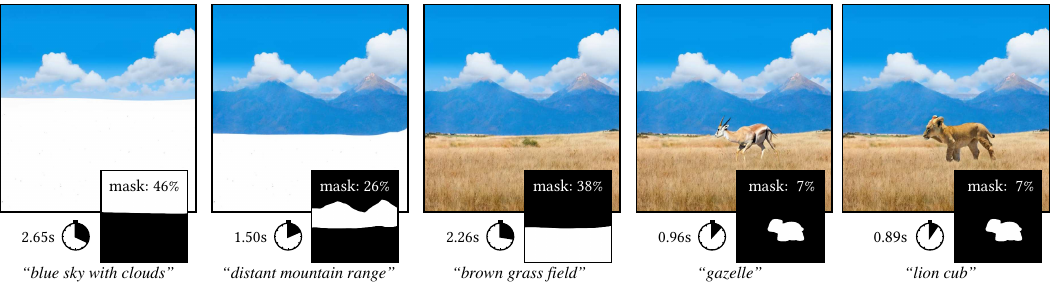} 
        \captionof{figure}{
            Incremental image generation at $1024\times1024$ using \methodname with 20 diffusion steps.
            The model generates content according to a text prompt in an area specified by a mask.
            Each update generates \emph{only} the masked pixels, with a runtime that depends chiefly on the size of the mask, rather than that of the image.
        }
        \label{fig:teaser}
    \end{center}
}]

\maketitle
\input{sections/0_abstract}    
\input{sections/1_intro}

\input{sections/2_rw}

\input{sections/3_method}
\input{sections/4_experiments}

\input{sections/6_conclusion}
{
    \small
    \bibliographystyle{ieeenat_fullname}
    \bibliography{main}
}

\newpage
\appendix
\input{sections/X_appendix}

\end{document}

%% file: packages.tex
\usepackage{xcolor}
\definecolor{cvprblue}{rgb}{0.21,0.49,0.74}

\usepackage{lipsum}

\usepackage{eccvabbrv}

\usepackage{graphicx}
\usepackage{booktabs}

\usepackage[accsupp]{axessibility}  %

\usepackage[pagebackref,breaklinks,colorlinks,citecolor=cvprblue]{hyperref}

\usepackage{orcidlink}

\usepackage{enumitem}
\usepackage[capitalize]{cleveref}
\usepackage{soul}
\usepackage{color, colortbl}
\usepackage{multirow}
\usepackage{adjustbox}
\usepackage{makecell}

%% file: macros.tex
\crefname{section}{Sec.}{Secs.}
\Crefname{section}{Section}{Sections}
\Crefname{table}{Table}{Tables}
\crefname{table}{Tab.}{Tabs.}

\setlist{nosep}

\definecolor{citecolor}{RGB}{30,102,235}

\makeatletter
\DeclareRobustCommand\onedot{\futurelet\@let@token\@onedot}
\def\@onedot{\ifx\@let@token.\else.\null\fi\xspace}

\def\eg{\emph{e.g}\onedot}

\def\etal{et~al\onedot}
\makeatother

\def\methodname{\emph{LazyDiffusion}\xspace}
\def\baselinecrop{\emph{RegenerateCrop}\xspace}
\def\baselinefull{\emph{RegenerateImage}\xspace}

\def\pixart{PixArt-$\alpha$\xspace}

\definecolor{mydarkblue}{rgb}{0,0.08,1}
\definecolor{mydarkgreen}{rgb}{0.02,0.6,0.02}
\definecolor{mydarkorange}{rgb}{1.0,0.5,0.02}
 \definecolor{mycyan}{rgb}{0.02,0.6,0.42}

\newif\ifdraft
\draftfalse

\ifdraft
    \newcommand{\ync}[1]{{\color{blue}\textbf{YN:} #1}}
    \newcommand{\dcc}[1]{{\color{red}\textbf{Danny:} #1}}
    \newcommand{\tae}[1]{{\color{orange}\textbf{Tae:} #1}}
    \newcommand{\mgc}[1]{{\color{mydarkgreen}\textbf{Michael:}#1}}
    \newcommand{\esc}[1]{{\color{mycyan}\textbf{ES:} #1}}
    \newcommand{\zwc}[1]{{\color{teal}\textbf{ZW:} #1}}
    
    \newcommand{\yn}[1]{{\color{blue}#1}}

\else
    \newcommand{\ync}[1]{}
    \newcommand{\dcc}[1]{}
    \newcommand{\mgc}[1]{}
    \newcommand{\MG}[1]{}
    \newcommand{\tae}[1]{}
    \newcommand{\zwc}[1]{}
    \newcommand{\esc}[1]{}
    
    \newcommand{\yn}[1]{{\color{black}#1}}

\fi

\newcommand{\myparagraph}[1]{\vspace{2mm} \noindent \textit{#1}}
\newcommand{\boldparagraph}[1]{\vspace{2mm} \noindent \textbf{#1}}

\newlist{todolist}{itemize}{2}
\setlist[todolist]{label=$\square$}

\definecolor{Gray}{gray}{0.9}

\newif\ifsingle
\singlefalse

%% file: sections/0_abstract.tex
\begin{abstract}

We introduce a novel diffusion transformer, \methodname, that generates
partial image updates efficiently.
Our approach targets interactive image editing applications in which, starting from a blank canvas or an image, a user specifies a sequence of localized image modifications using binary masks and text prompts.
Our generator operates in two phases. 
First, a context encoder processes the current canvas and user mask to produce a compact global context tailored to the region to generate.
Second, conditioned on this context, a diffusion-based transformer decoder synthesizes the masked pixels in a ``lazy'' fashion, i.e., it \emph{only} generates the masked region.
This contrasts with previous works that either regenerate the full canvas, wasting time and computation, or confine processing to a tight rectangular crop around the mask, ignoring the global image context altogether.
Our decoder's runtime scales with the mask size, which is typically small, while our encoder introduces negligible overhead.
We demonstrate that our approach is competitive with state-of-the-art inpainting methods in terms of quality and fidelity while providing a $10\times$ speedup for typical user interactions, where the editing mask represents 10\% of the image.

\end{abstract}

%% file: sections/1_intro.tex
\vspace{-3mm}
\section{Introduction}
\label{sec:intro}

Diffusion models have had remarkable successes in generating high-quality and diverse images.
They are the powerful engine behind exciting local image editing applications based on inpainting, where a user provides a mask and a text prompt describing a region to modify and the content to generate, respectively~\cite{rombach2022high,wang2023imagen,nichol2021glide}.
While current approaches yield impressive results, they are also slow and wasteful.
Invisible to the end user, the inpainting pipeline generates an entire image and then selectively utilizes only the few pixels located within the mask, discarding all others.
Although this approach is generally common in inpainting pipelines~\cite{zhao2021comodgan,yu2019free}, its inefficiency is particularly pronounced with diffusion models, due to their iterative sampling procedure, precluding their usage in interactive workflows.
Practitioners~\cite{sd-webui,diffusers} save time and computation by cropping a small rectangular region around the mask, possibly downsampling for processing with the diffusion, then upsampling and blending the result to fill the hole.
In doing so they compromise image quality and sacrifice the global image context, which often leads to spatially inconsistent outputs  
(Compare Figs.~\ref{fig:local-edits-comparison}(a) and~\ref{fig:local-edits-comparison}(b)).

We propose a new generative model architecture, which we call \methodname.
Our approach, illustrated in Fig.~\ref{fig:teaser}, generates \emph{partial} image updates, strictly limited to the masked region, and does so efficiently, with a cost commensurate to the mask size.
Yet, its output respects the global context given by the observed canvas (Fig.~\ref{fig:local-edits-comparison}(c)).
To achieve this, our key idea is to decouple the generative process into two distinct steps.
First, an encoder processes the visible canvas and mask, summarizing them into a global context code.
This encoder processes the entire canvas, but it only runs once per mask.
Second, conditioned on the global context and the user's text prompt, a diffusion decoder generates the next partial canvas update.
This model runs many times during the diffusion process, but unlike previous works, it only operates on the masked region.
Since, in practice, most updates cover small areas (10--20\% of the image), this yields significant computation savings, thus making the editing experience more interactive.
\input{figures/local_edits_comparison}

Our encoder and diffusion decoder operate in a latent space~\cite{rombach2022high}, for efficiency.
Both use the transformer architecture~\cite{vaswani2017attention,dosovitskiy2020image,peebles2023scalable}.
The transformer architecture is particularly appealing 
because splitting the image into small enough patches (tokens) enables generating arbitrarily-shaped regions with minimal waste.
The encoder processes the entire image and mask and produces a mask-dependent context.
We keep only the context tokens corresponding to the location of the masked patches.
This ensures the downstream computation only scales with the size of the masked region, and encourages the compressed context to represent the relationship of the masked region to the rest of the image.
At each denoising step, the decoder only processes tokens corresponding to masked patches.
While the decoder \emph{generates} only the masked region, it \emph{``sees''} the entire image, through the compressed context, ensuring strong coherence.
The conditioning on context is efficient and adds negligible computational overhead.
In contrast, previous methods~\cite{avrahami2022blended,wang2023imagen,rombach2022high} achieve spatial  consistency by uniformly processing all image regions, masked or not.
Figure~\ref{fig:compare-to-full} illustrates the conceptual difference between our approach and a baseline diffusion transformer.

Our approach reduces computational cost significantly for small masks, typical in interactive editing.
We achieve a speedup up to $\times10$ over methods processing the entire image, for mask covering 10\% of the image.
Additionally, our model produces results of comparable quality, indicating that the compressed context is rich and expressive.
In an interactive image generation context, our method amortizes the overall synthesis cost over multiple user interactions, improving interaction latency.
It also amortizes the encoder cost when generating multiple updates for a given mask, using different input noise or text prompt (\cref{fig:teaser}, rightmost panel).
\input{figures/compare_to_full}

%% file: figures/local_edits_comparison.tex
\begin{figure}
    \centering
    \includegraphics[width=\linewidth]{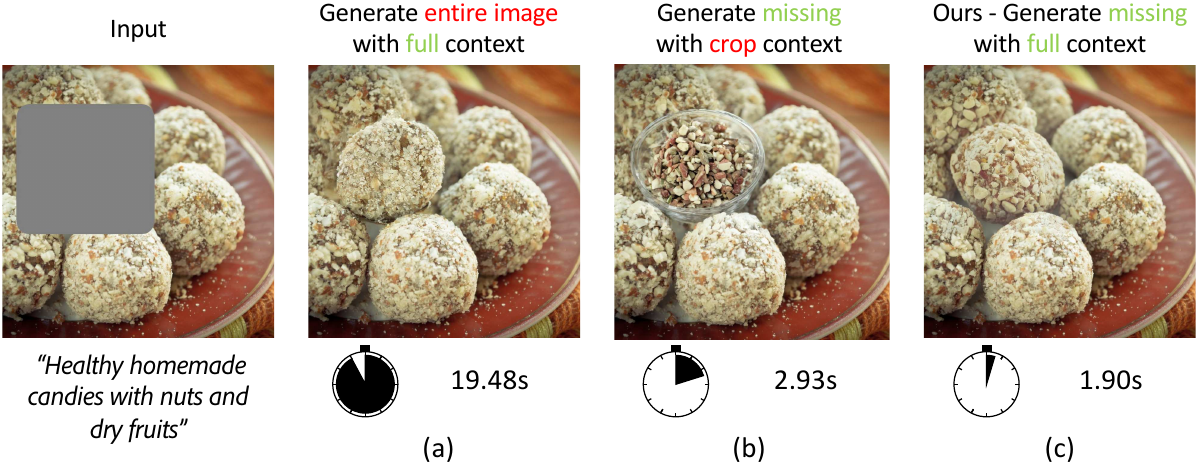}
    \caption{\label{fig:local-edits-comparison}
    Comparing inpainting approaches.
    (a) Most works~\cite{rombach2022high,podell2023sdxl} generate the entire image, utilizing the full image context and fill the hole by discarding the non-masked regions.
    While the outcome aligns well with the image, the process is time-consuming.
    (b) generating only a lower resolution crop around the mask is more efficient and still seamlessly blends with nearby pixels~\cite{sd-webui,diffusers}.
    However, the inpainted content is semantically inconsistent with the overall image context.
    (c) our approach ensures both global consistency and efficient execution.
}
\end{figure}

%% file: figures/compare_to_full.tex
\begin{figure}[t]
    \centering
    \ifsingle
        \includegraphics[width=\linewidth]{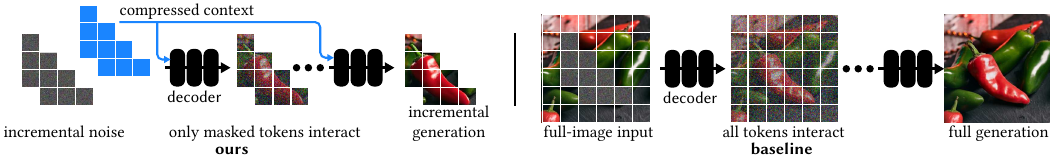}
    \else
        \includegraphics[width=\linewidth]{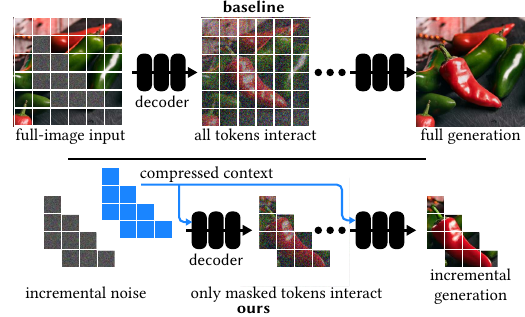}
    \fi
    \vspace{-5mm}
    \caption{
    \label{fig:compare-to-full}
        Our diffusion transformer decoder 
        \ifsingle (left) \else (bottom) \fi
        reduces synthesis computation using two strategies.
        First, we compress the image context using a separate encoder (not shown) outside the diffusion loop.
        Second, we only generate tokens corresponding to the masked region to generate.
        In contrast, typical diffusion transformers 
        \ifsingle (right) \else (top) \fi ~\cite{peebles2023scalable,chen2023pixartalpha} maintain tokens for the entire image throughout the diffusion process, to preserve global context.
        When performing inpainting, such model generates a full-size image, most of which is discarded in order to in-fill the hole region only.
        Existing convolutional diffusion models for inpainting~\cite{rombach2022high} suffer from the same drawbacks.
    }
\end{figure}

%% file: sections/2_rw.tex
\section{Related Work}
\label{sec:rw}

\myparagraph{Speeding up diffusion models.}
Diffusion models~\cite{ho2020denoising,song2020score,sohl2015deep} are a significant breakthrough in generative modeling~\cite{dhariwal2021diffusion,rombach2022high,saharia2205photorealistic,ramesh2022hierarchical,betker2023improving} and editing~\cite{meng2021sdedit,avrahami2022blended}, producing images with unparalleled quality and diversity.
But they remain costly to evaluate, due to the iterative nature of their sampling process.
Numerous methods have been developed to improve their inference time, such as better samplers and dedicated ODE solvers~\cite{song2020denoising,karras2022elucidating,lu2022dpm,lu2022dpm2}, distillation techniques~\cite{liu2023instaflow,song2023consistency,salimans2022progressive,luo2023latent}.
The gap between recent one-step diffusion models~\cite{yin2023one,podell2023sdxl,nguyen2023swiftbrush} and their expensive multi-step counterparts is closing.
Our approach also seeks to speed up the image synthesis process for diffusion-based models, but our contribution is largely orthogonal and can be combined with these optimizations: we reduce the amount of image data to process, rather than the atomic diffusion iteration.

\myparagraph{Transformer-based generative models.}
Early transformers for image generation generate image autoregressively~\cite{razavi2019generating,esser2021taming,chen2020generative} in scanline order.
CogView2~\cite{ding2022cogview2} proposes a hierarchical transformer to improve generation speed and shows application to text-guided image inpainting with rectangular masks.
Later non-autoregressive models like MaskGIT~\cite{chang2022maskgit} generate images gradually, a few tokens at a time, but they do so iteratively, generating all tokens at every iteration and discarding the unmasked ones, which is inefficient.
They focus on sequential generation to improve image quality.

Our transformer-based model design is inspired by Masked Autoencoders (MAEs)~\cite{he2022masked}, but we reverse their asymmetric design.
Our encoder processes \emph{all} the tokens to produce context at the masked locations, and our decoder operates on the masked tokens.
Our decoder is a powerful diffusion transformer, recently proposed as an alternative to the popular UNet design~\cite{tevet2022human,peebles2023scalable}.
Most relevant to this work, DiT~\cite{peebles2023scalable} was proposed for class-conditioned image generation and was improved in \pixart~\cite{chen2023pixartalpha} to support text-conditioning.
Our diffusion decoder is an adaptation of \pixart that additionally conditions on the global context produced by the encoder.
Masked diffusion transformers were previously explored for representation learning~\cite{gao2023masked,wei2023diffusion} or for minimizing training cost~\cite{zheng2023fast}.
Our focus is on speeding inference to improve interactivity.
Recent trends indicate that the transformer architecture becoming central to state-of-the-start image~\cite{esser2024scaling} and video generators~\cite{videoworldsimulators2024}, for which our method would enable faster inference and interactive applications.

\myparagraph{Text-guided diffusion-based image editing.}
Text-to-image diffusion models have become the de-facto foundation for generative image editing methods.
With user edits typically spatially localized, significant effort has gone into developing techniques that allow precise modifications~\cite{hertz2022prompt,cao2023masactrl,patashnik2023localizing} by selectively manipulating internal representations, \eg attention maps, during the denoising process to affect only certain local regions without undesirable side-effects.
Another line of work adopts the formulation of inpainting, where a mask is provided to localize the edit.
Blended diffusion~\cite{avrahami2022blended} and DiffEdit~\cite{couairon2022diffedit} use pre-trained generation models and spatially blend noised versions of the input into the gradual denoising process to enforce the preservation of unmasked regions.
This indirect approach often result in artifacts, leading more recent approaches to fine-tune text-to-image models specifically for inpainting.
Starting from an image generation architecture, GLIDE~\cite{nichol2021glide} and Stable Diffusion Inpaint~\cite{rombach2022high} add mechanisms to additionally condition on the mask and masked image and fine-tune the models to predict the masked pixels.
Recent advancements in this domain involve training inpainting models with object-level masks~\cite{wang2023imagen} rather than random ones and possibly also object-level text captions~\cite{xie2023smartbrush}, mirroring real-world usage more closely.
These works retrofit image generation architectures for local editing,
but these models produce the full image, including regions that should not be changed.
This is inefficient in time and computing resources.
Our architecture efficiently performs local edits by generating only the masked region.

%% file: sections/3_method.tex
\section{Method}
\label{sec:method}

Our goal is to develop an efficient diffusion generator for text-guided image editing, whose generation cost scales with the size of the 
region to generate, and which can incorporate the context of the entire image for a fixed, small fraction of its total cost.
Starting from an image $I\in \mathbb{R}^{h\times w\times 3}$, the user specifies the region to be edited with a binary mask $M \in \{0, 1\}^{h\times w}$ and text prompt $\mathbf{c}$, indicating where and what content to generate. A mask value $1$ specifies a hole to inpaint, and $0$ for context pixels to not touch. Unless stated otherwise, we use images of $h=w=1024$ resolution.

Following standard practice, we operate in latent space~\cite{rombach2022high}, a compressed version of the RGB domain (\S~\ref{subsec:latent}). Observing that the iterative diffusion process is the computational bottleneck in state-of-the-art generators, our generator has a novel asymmetric encoder-decoder transformer architecture, as illustrated in Fig.~\ref{fig:coarse_arch}.
The encoder (\S~\ref{subsec:context-encoder}) compresses and summarizes the whole image context and is only run once.
The decoder (\S~\ref{subsec:local-decoder}) is a transformer-based diffusion denoiser that is iteratively run, but only on the masked area.
As such, computation cost and latency are proportional to the number of pixels to synthesize, rather than the entire canvas~\cite{xie2023smartbrush,wang2023imagen,avrahami2022blended}.
This significantly reduces computation since, for most edits, the masks are small.

\input{figures/overview}
\subsection{Latent space processing}\label{subsec:latent}

Following previous works of Latent Diffusion Models (LDM)~\cite{rombach2022high}, our model operates
in an intermediate latent space of $8\times$ lower resolution with $c=4$ channels, which reduces computation without significantly impacting visual quality.
We use the pretrained latent VAE of Stable Diffusion~\cite{rombach2022high}, denoting the encoder and decoder $\mathcal{E}$ and $\mathcal{D}$, respectively.
We encode the masked image as our latent input~\cite{wang2023imagen}:
\begin{equation}
\begin{split}
Z = \mathcal{E}\left(I \odot (1-M) \right) \hspace{2mm} \in\mathbb{R}^{\frac{h}{8}\times \frac{w}{8} \times c},    
\end{split}  
\end{equation}
where $\odot$ represents element-wise multiplication across the spatial dimensions. 

\subsection{Global context encoder}
\label{subsec:context-encoder}

Encoder $E$ processes the whole image, with the goal of efficiently encoding the information given by the visible region, so that a downstream decoder can synthesize a visually consistent output with the context.
Our encoder $E$ is a Vision Transformers (ViT)~\cite{dosovitskiy2020image}.
To produce tokens, we first downsample the mask $M$ using a learned convolution layer to match the latent spatial dimensions, as done by Wang et al.~\cite{wang2023imagen}. Then, we concatenate the downsampled mask and latent code $Z$ along the channel dimensions and and divide them into $4\times4$ patches, with an overlap of $1$ on each side. 
This yields $N = 64 \times 64 = 4096$ patches.
Then, following standard practice, we linearly embed each patch and add positional embedding~\cite{vaswani2017attention}.
Finally, the tokens are passed through the transformers and produce a new set of $N$ tokens.
In summary,
the encoder transforms the input $Z$ and $M$ into a set of $N$ tokens of dimension $d=1152$.

\begin{equation}\label{eq:global-context}
    \mathcal{T}_\text{all} = \{\mathbf{\tau}_1, \mathbf{\tau}_2, \ldots, \mathbf{\tau}_N \} = E(Z, M), ~\mathbf{\tau}_i\in\mathbb{R}^d.
\end{equation}

\myparagraph{Token dropping.} The set of output tokens contain information regarding the whole image, but using them all would cause downstream computation to scale with respect to the input size. \emph{Can we instead keep only a subset of tokens, that would hold the information needed for generation?}

\ync{work on this, unclear to me.}
As the self-attention layers in the encoder transformer enable all the tokens to interact, each individual token has the potential to encode the relevant context of the whole image.
As such, we discard the tokens corresponding to the visible region, keeping the ones corresponding to the hole.
Dropping tokens outside the mask creates an information bottleneck that encourages $E$ to summarize the input context in a compact set of tokens and ensures the downstream computation only scales with the size of the masked area, since the decoder will thus only process tokens covering the hole. 
The tokens should also represent the relevant information for the given location; previous works visualizing transformers~\cite{caron2021emerging} suggest that this location information can be preserved.
Patches with partial holes are also included, and the visible pixels in those patches are blended in at the output step. Formally, we maxpool mask $M$ to a $64\times 64$ map and vectorize into a set $\{m_i\}_{i=1}^{4096}$, where $m_i\in \{0, 1\}$.

\begin{equation}\label{eq:global-context}
    \mathcal{T}_\text{hole} = \{\mathbf{\tau}_i \ | \hspace{1mm} m_i = 1\}
    \subseteq \mathcal{T}_\text{all}.
\end{equation}

\noindent The remaining set of $N_\text{hole}\leq N$ tokens form our compressed global context. 
This design, along other architectural choices, are evaluated in the supplemental.

\subsection{Incremental diffusion decoder}\label{subsec:local-decoder}

We synthesize the missing pixels, using a transformer-based diffusion decoder $D$~\cite{chen2023pixartalpha,peebles2023scalable}. 
Rather than keeping a set of $N$ tokens
representing the whole image, we start with $N_\text{hole}$ tokens corresponding to the hole,
$\mathcal{X}_\text{hole}=\{\mathbf{x}_i\}$.
The diffusion process creates time-conditioned tokens $\mathcal{X}_\text{hole}^t=\{\mathbf{x}_i^t\}$, where $t\in [0, ..., T]$, starting at time $T$ with features drawn from a unit Gaussian.
The decoder progressively denoises these tokens, conditioned on the T5-encoded text prompt $\mathbf{c}$~\cite{raffel2020exploring} and the global context produced by the encoder $\mathcal{T}_\text{hole}$: 

\begin{equation}\label{eq:diffusion}
    \mathcal{X}_\text{hole}^{t-1} =  D\left( \mathcal{X}_\text{hole}^{t} \oplus \mathcal{T}_\text{hole};
    t, \mathbf{c}\right),
\end{equation}
where $\oplus$ denotes concatenation along the hidden dimension
of corresponding elements in each set.
We find this conditioning mechanism superior to several alternatives analyzed in~\cref{sec:imagenet}.

\myparagraph{Blending.} The final tokens $\mathcal{X}^0_\text{hole}$ are mapped back into the latent image domain using a linear layer, and the inverse of the patch-splitting procedure to obtain a partial latent image $\hat{Z}_\text{hole}\in \mathbb{R}^{\frac{h}{8}\times \frac{w}{8} \times c}$.
The missing tokens, corresponding to visible pixels, are left uninitialized with zeros.
We combine this output with the visible latent, using pointwise masking, to obtain the final latent composite:
\begin{equation}\label{eq:output-latent}
    \hat{Z} = (1-M) \odot Z + M \odot \hat{Z}_\text{hole}.
\end{equation}
Finally, this is decoded by the latent decoder to produce the final RGB image $\hat{I} = \mathcal{D}(\hat{Z})$.

These decoded results occasionally contains faintly visible seams. 
Previous works performing inpainting with latent diffusion models observed this phenomenon and addressed it with a dedicated latent decoder~\cite{zhu2023asymmetric}.
As their decoder is computationally intensive, we opt to use a simple Poisson blending postprocessing step~\cite{perez2003poisson} in RGB space.
We discuss this challenge in greater length in the supplemental.

\myparagraph{Training and implementation details.}
For the decoder, we adopt the \pixart~\cite{chen2023pixartalpha} architecture, and add a single layer to support our conditioning on context.
We initialize all shared layers from the public \pixart checkpoint to benefit from their pretraining.
The encoder on the other hand, is trained from scratch.
The two models are trained jointly to reconstruct masked (latent) pixels, using the Improved DDPM objective~\cite{nichol2021improved}. 
We train our model for 100,000 iterations on 56 NVIDIA A100 GPUs, using the AdamW optimizer~\cite{loshchilov2017decoupled}, with a constant learning rate $2\times 10^{-5}$, weight decay set to $3\times 10^{-2}$ and global batch size of 224.
We use $T=1000$ diffusion steps during training.
We generate our results using the Improved DDPM sampler~\cite{nichol2021improved} with 50 steps, unless specified otherwise, and set the classifier-free guidance scale to 4.5.
All running times are measured on a single A100 GPU.
We provide further details in~\cref{sec:details}.

%% file: figures/overview.tex
\begin{figure*}[t]
    \centering
    \includegraphics[width=\linewidth]{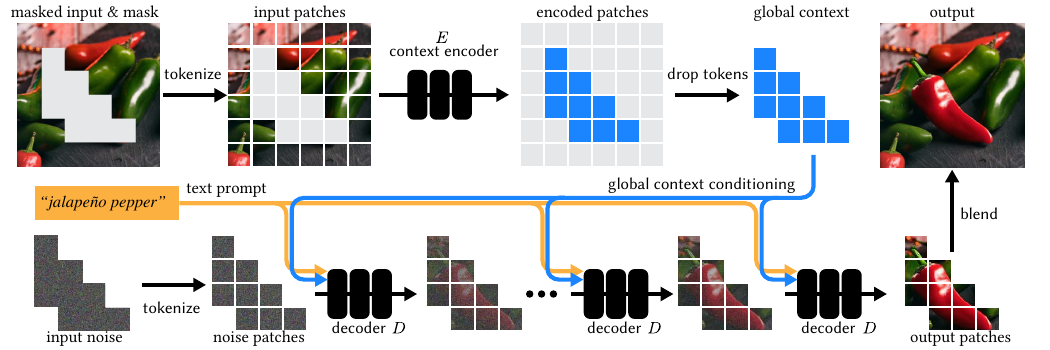}
    \caption{
    {\bf Overview.} 
    To generate an incremental image update, our algorithm takes as input a user mask and a text prompt.
    {\bf(top)} We start by transforming the visible pixels and binary mask into patches, 
    and pass them to a vision transformer (ViT) encoder.
    We then drop all tokens, except those corresponding to the hole region; this is our global context.
    {\bf (bottom)} To generate the missing pixels, we initialize
    a set of noise patches corresponding to the masked region and pass them through a diffusion transformer model for several denoising iterations, until we obtain denoised patches.
    Unlike previous works~\cite{peebles2023scalable,chen2023pixartalpha}, which process the entire image, our diffusion transformer only processes the patches required to cover the missing region. 
    We train our encoder and diffusion decoder jointly using a diffusion denoising objective on the missing patches.
    The generated patches are then blended back into the missing region to produce the final output. 
    Our model operates in a pretrained latent image space~\cite{rombach2022high},
    but we illustrate our pipeline with RGB images for simplicity.
    } 
    \label{fig:coarse_arch}
\end{figure*}

%% file: sections/4_experiments.tex
\section{Experiments}
\label{sec:experiments}

\subsection{Experimental setup}
\label{subsec:setup}

\yn{
The main paper primarily focuses on a text-conditioned setting, as do the experiments that follow.
However, our approach is versatile and can be applied in other use cases as well.
In the early stages of this research, we primarily explored unconditional inpainting on the ImageNet dataset~\cite{deng2009imagenet}, which are detailed in~\cref{sec:imagenet}.
}

\myparagraph{Dataset.}
We train our model at $1024\times1024$ resolution on an internal dataset containing 220 million high-quality images, covering a wide variety of objects and scenes.
We produce masks and text prompts in a process similar to that proposed by Xie et al. \cite{xie2023smartbrush}.
Specifically, we use an entity segmentation model~\cite{qi2022open} to segment all objects in an image and then caption each entity with BLIP-2~\cite{li2023blip}.
To simulate the rough and inaccurate masks created by users, we randomly dilate the entity mask (see ~\cref{sec:details} for details).
During training, we randomly sample triplets of image, mask, and caption.

\myparagraph{Baselines.}
We compare \methodname with two inpainting baselines (already shown in~\cref{fig:local-edits-comparison}), which we refer to as \baselinefull and \baselinecrop.
\baselinefull, is the approach found in most academic works~\cite{wang2023imagen,xie2023smartbrush,rombach2022high,podell2023sdxl}, and operates on the entire image.
\baselinecrop, used in popular software frameworks~\cite{sd-webui,diffusers}, operates on a tight square crop around the masked region.
The crop is first resized to a fixed low-resolution before processing and is upsampled back afterwards.
Both approaches generate as many pixels as their 
input contains (whether full-canvas or local crop), unlike \methodname that generates only masked patches.

To ensure a fair comparison, we utilize the \pixart architecture for both approaches.
Since there is currently no publicly available PixArt-based inpainting models, we design and train them ourselves.
We adapt PixArt for inpainting using the same procedure employed to transform Stable Diffusion~\cite{rombach2022high} from generation to inpainting.
Specifically, we incorporate the GLIDE~\cite{nichol2021glide} conditioning mechanism, where the generator operates on 9 latent channels: four channels for the latent being denoised, four channels representing the latent of the masked input image, and the last channel containing a downsampled version of the mask.
We train two PixArt models at $1024\times1024$ and $512\times512$ for \baselinefull and \baselinecrop, respectively.

We also compare with Stable Diffusion variants of these two approaches for reference: SDXL~\cite{podell2023sdxl} operates on the entire $1024\times1024$ image, while SD2-crop~\cite{rombach2022high} operates on a $512\times512$ crop.
It is important to note that these models utilize different architectures and were trained on different datasets, and hence are not directly comparable.
We include them in this comparison only as references for state-of-the-art quality.

\input{figures/runtime}

\subsection{Inference time}
We illustrate the overall runtime of all methods in \cref{fig:runtime}.
The baselines run is constant time, as they operate on fixed size tensors derived from the fixed input size -- full canvas for \baselinefull and a fixed-size crop for \baselinecrop.
In contrast, \methodname's runtime scales with the mask size, because our decoder processes tensors with dimensions proportional to the masked region.
This leads to significant speedups for small masks, typical of interactive editing applications.
For example, with a mask covering 10\% of the image our model achieves a $\times10$ speedup over \baselinefull.
Similarly, \methodname is also faster than \baselinecrop for masks smaller than 25\%.
At mask ratio 25\%, both methods generate the same number of pixels and have comparable running times.
For larger masks, \baselinecrop is faster but generates low-resolution crops and naively upsamples to native resolution, reducing sharpness. 
Additionally, \baselinecrop often fails to produce outputs that are consistent with the region outside the mask, as we discuss below (\cref{subsec:quality}).

While there are additional networks in the pipeline, the diffusion decoder is the only component running multiple times, and thus dominates the runtime.
Notably, our context encoder adds a $73$ms overhead, which is dwarfed by the cost of the diffusion loop.
The latent encoder and decoder take $97$ms and $176$ms, respectively, and the T5 text encoder $21$ms. 
These are shared by all methods.

\myparagraph{Scaling laws.} 
Our method essentially reduces the cost of each denoising iteration at the price of a small overhead for the context encoder, to balance quality with context retention.
As a result, our performance gains are most striking for high diffusion step counts (typically correlated with higher image quality), and smaller mask sizes (most frequent in interactive applications).
A single evaluation of our decoder takes $374$ms to generate \emph{full} image,
but only $28$ms for 10\% masks --- a $\times13.4$ speedup, greater than the encoder's overhead.
So, our method remains beneficial for few-step~\cite{song2023consistency}, or even one-step models~\cite{yin2023one}.
We expect the performance gains provided by our strategy to be even more striking on costlier applications like high-resolution image editing, or video synthesis~\cite{videoworldsimulators2024}.

\subsection{Progressive generation}

Diffusion models are challenging to integrate into interactive pipelines due to their high latency.
\input{figures/gradual_results}
There exists an abundance of research on broadly accelerating diffusion models~\cite{yin2023one,song2023consistency,lu2022dpm}, but in the context of this study, we highlight that individuals often tackle tasks incrementally, executing operations progressively and concentrating on local adjustments one at a time—whether it involves adding or removing objects, refining, or retrying previous attempts. 
\methodname significantly accelerates such local operations, making it well-suited for interactive pipelines with a user-in-the-loop.

In~\cref{fig:gradual_generation}, we showcase a couple of iterations using \methodname for both image editing and image generation, starting from a blank canvas.
Furthermore, we attach a supplemental video that showcases authentic user interactions with both \methodname and our \baselinefull baseline, highlighting the discernible difference in running time between the two.

\subsection{Inpainting quality}
\label{subsec:quality}

A distinctive feature of \methodname is its utilization of a compressed global context to aid inpainting.
In contrast, \baselinefull utilizes the complete global context, while \baselinecrop relies on the context provided by pixels neighboring the mask.
We now compare the results produced by these approaches.

For quantitative evaluation, we report zero-shot FID~\cite{heusel2017gans} and CLIPScore~\cite{hessel2021clipscore}, which estimate similarity to real images and text-image alignment, respectively.
Additionally, we include scores for SDXL~\cite{podell2023sdxl} and SD2-crop~\cite{rombach2022high}.
Despite not being directly comparable, because they use different architectures and training data, they serve as references for state-of-the-art quality.
In Table~\ref{tab:quant}, we report mean scores over a random sample of 10,000 images drawn from OpenImages~\cite{schuhmann2021laion}.
Notably, text-image alignment (CLIP) remains unaffected by the mechanism to use image context.
On the FID metric, \methodname exhibits only a marginal increase compared to \baselinefull (\%4) and performs significantly better than \baselinecrop (\%26).
\input{figures/quant}

\input{figures/qual_compare}

We show qualitative comparisons in \cref{fig:qual_compare}.
Our examination reveals a significant discrepancy in the performance of models regenerating a crop -- \baselinecrop and SD2-crop.
In many instances, inpainting involves generating an object that is visually independent of other concepts in the image, such as adding a side of fries next to a burger.
Here, models operating on a tight crop can produce reasonable-looking objects and seamlessly blend them with the surrounding pixels available in the crop (\cref{fig:qual_compare} (Top)).
However, in numerous scenarios, the goal is to add an object that is strongly related to the existing context, such as adding another bun to a tray of buns. 
Models operating solely on a crop lack knowledge of the global image and consequently produce objects that may seem reasonable in isolation but do not fit well within the greater image context (\cref{fig:qual_compare} (Bottom)).
In contrast, SDXL and \baselinefull utilize direct and full access to all image pixels to consistently yield highly realistic results, where the generated region fits well with the existing content. 
Notably, we find that \methodname behaves similarly and produces comparable results even in these challenging edge cases.
This suggests that the compressed image context is highly expressive and encodes meaningful semantic information.

\myparagraph{User study.}
We measure the models' capability to produce highly-contextual inpainting through a user study.
For this, we curate a specialized test set comprising scenarios that necessitate a high level of semantic image context for effective inpainting. 
Specifically, we select images featuring several closely related objects, such as a set of uniform buns on a tray. 
Subsequently, we evaluate all models based on their ability to regenerate one of these objects when masked. 
In this scenario, the models must rely on visible pixels to produce a high-fidelity result. 
Users are presented with the masked input image, a text prompt, and two results — ours and a baseline. They are then asked to "select the option in which the inpainted image, as a whole, looks best".
We collect a total of 1778 responses from 48 unique users and find that our method is strongly preferred over methods operating solely on a crop and competitive with those regenerating the entire image.
Specifically, \methodname is preferred over \baselinecrop in 81\% of cases, over SD2-crop in 82.5\% of cases, over \baselinefull in 46.1\% of cases, and over SDXL in 48.5\% of cases.
These results indicate that the compressed encoder context retains the core semantic information required even for challenging use cases.
In short, our model demonstrates competitive quality to our conceptual upper-bound \baselinefull, but runs up to ten times faster.

\input{figures/SDEdit}
\subsection{Sketch-guided inpainting}
So far, our emphasis has been on generation guided solely by the mask and a text prompt. 
However, in principle, our method is applicable to any localized generation task and can accommodate other forms of conditioning, such as sketches and edge maps. 
In~\cref{fig:sdedit}, we briefly showcase this versatility by guiding the generation with a coarse color sketch provided by the user. 
Following the SDEdit~\cite{meng2021sdedit} approach, we initiate the generation process from the partially noised input image instead of Gaussian noise.

%% file: figures/runtime.tex
\begin{figure}[t]
    \centering
    \ifsingle
        \includegraphics[width=\linewidth]{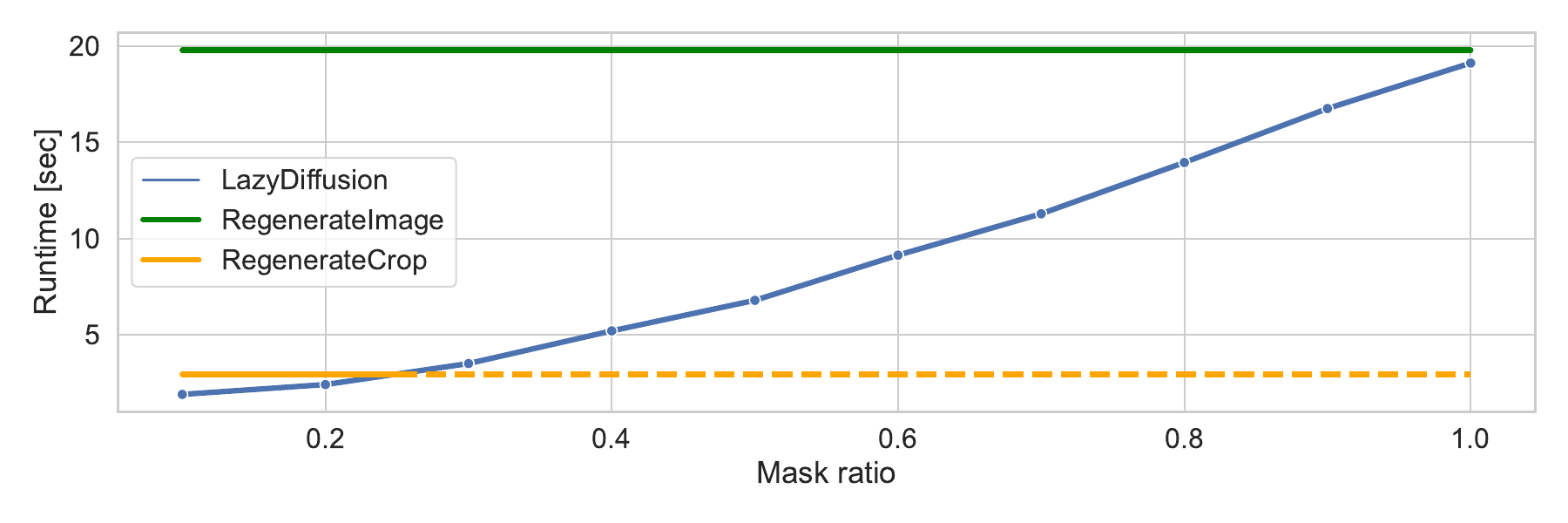}
    \else
        \includegraphics[width=\linewidth]{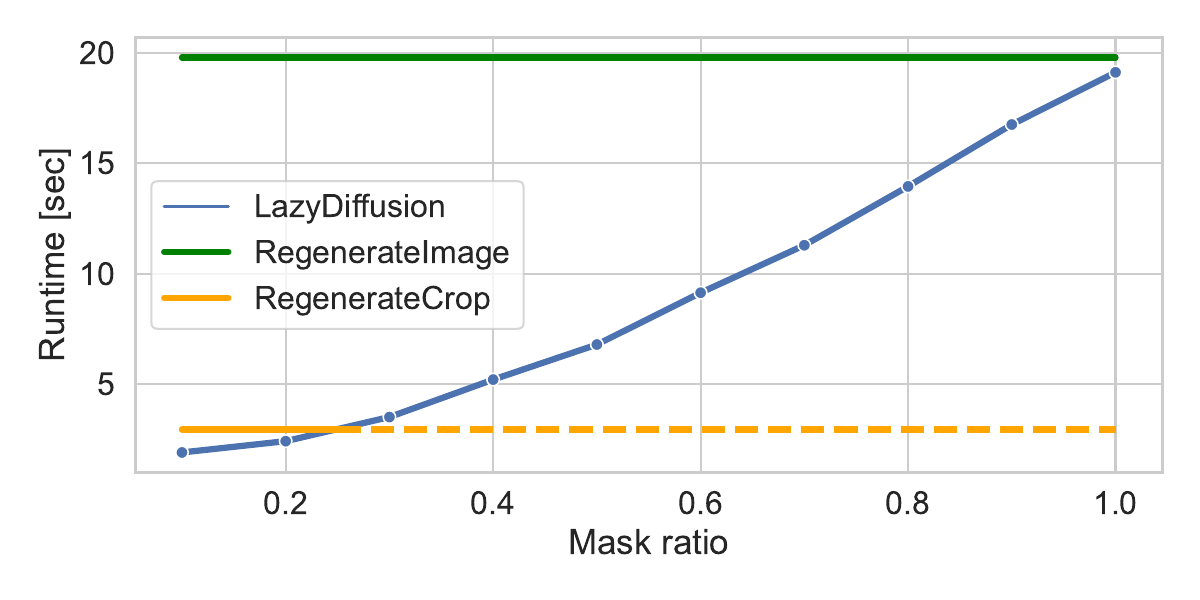}
    \fi
    
    \caption{
        Comparing \methodname's runtime to that of baselines regenerating the entire $1024\times1024$ image or a smaller $512\times512$ crop around the mask.
        \methodname is consistently faster than \baselinefull, especially for small mask ratios typical to interactive edits, reaching a speedup of $10\times$.
        Similarly, \methodname is faster than \baselinecrop for mask ratios $<25\%$. For masks greater than that (dashed), \baselinecrop is technically faster but generates in low-resolution and naively upsamples to match the desired resolution, harming image quality.
    }
    \label{fig:runtime}
    \vspace{-5mm}
\end{figure}

%% file: figures/gradual_results.tex
\ifsingle
    \begin{figure}[t]
        \centering
        \begin{tabular}{cc}
             \hspace{-1cm} \includegraphics[width=0.55\linewidth]{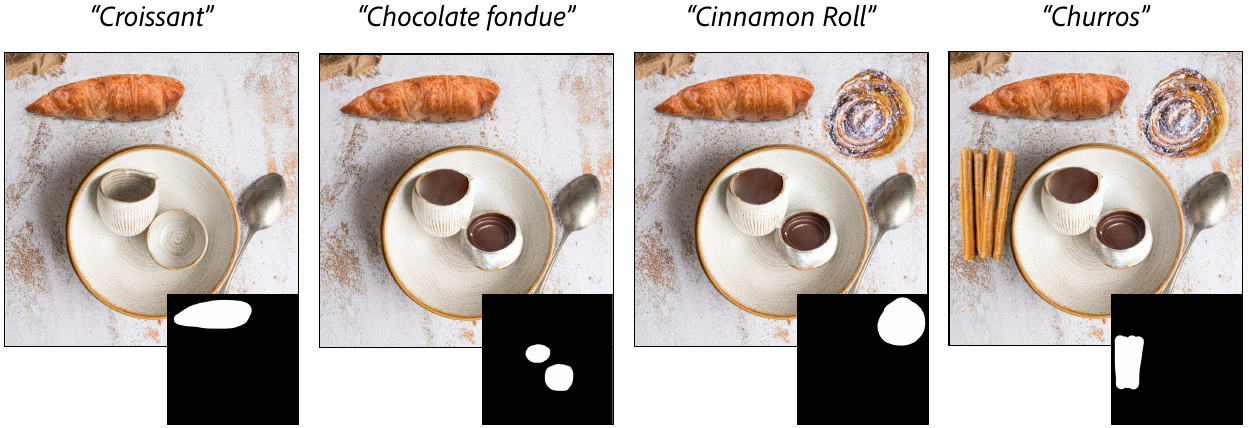} &
             \hspace{1mm} \includegraphics[width=0.55\linewidth]{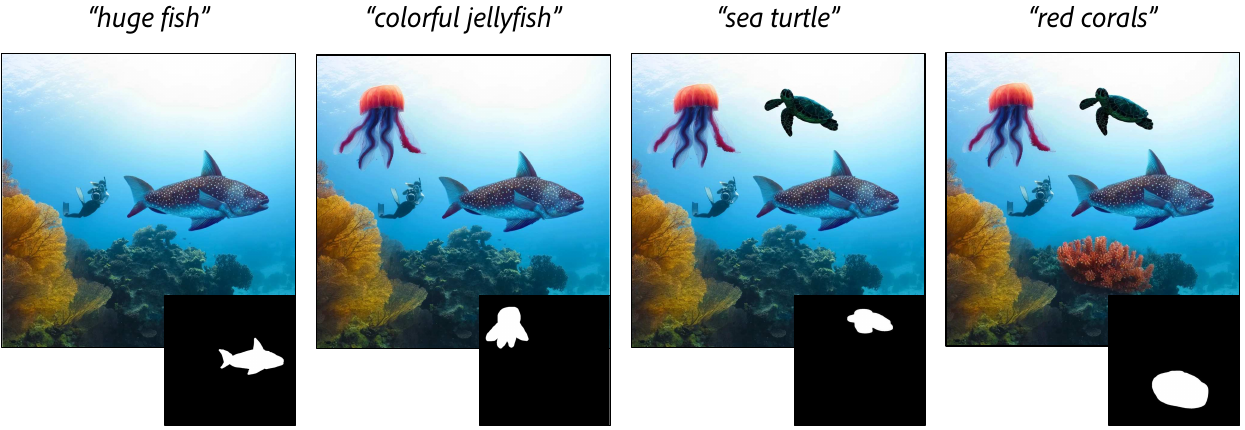} \\
             \midrule
             \hspace{-1cm} \includegraphics[width=0.55\linewidth]{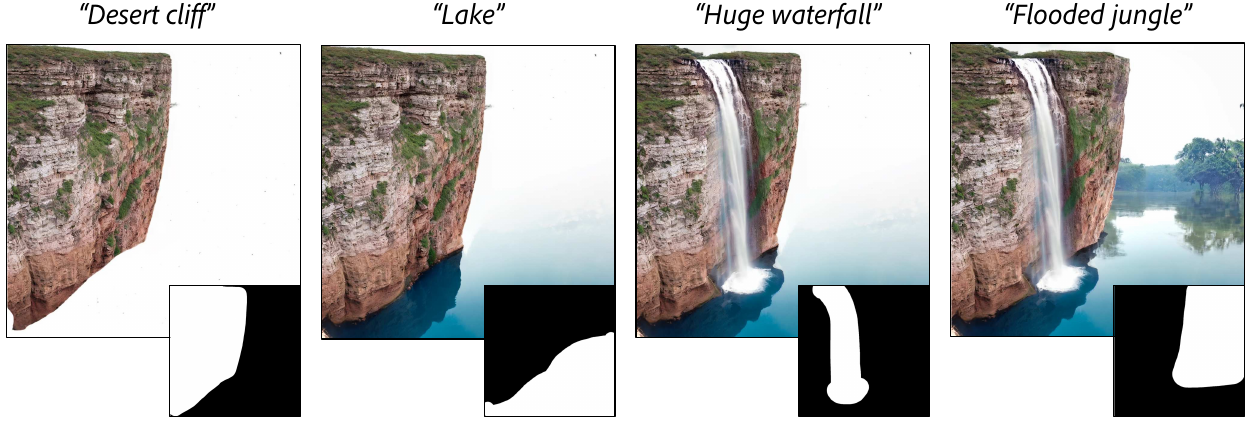} &
             \hspace{1mm} \includegraphics[width=0.55\linewidth]{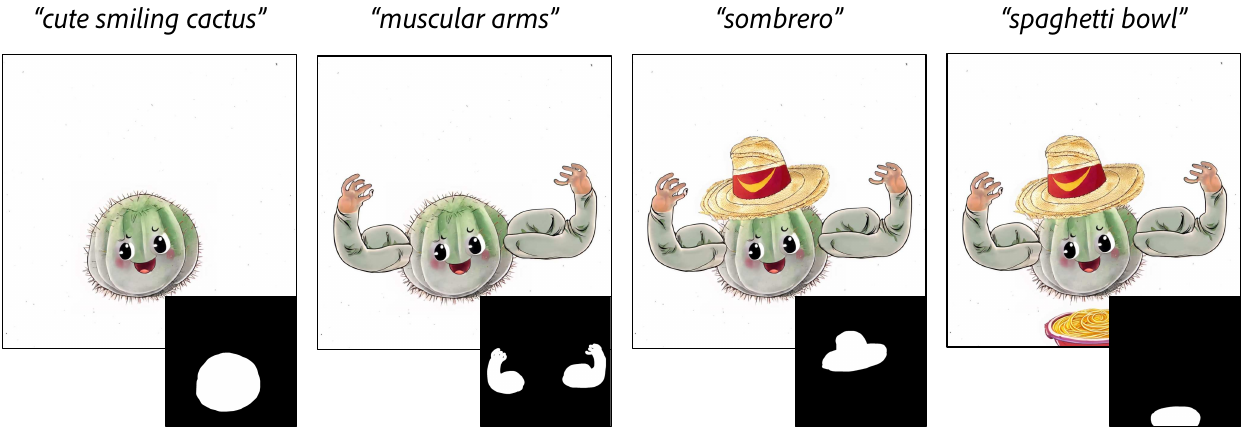} \\
        \end{tabular}
        \caption{
        Progressive image editing (top) and image generation (bottom) using \methodname.
        Each panel illustrates a generative progression compared to the preceding state of the canvas to its left.
        \methodname markedly accelerates local image edits (approximately $\times10$), rendering diffusion models more apt for user-in-the-loop applications.
        }
        \label{fig:gradual_generation}
        \vspace{-5mm}
    \end{figure}
\else
    \begin{figure}[t]
        \centering
        \begin{tabular}{c}
             \includegraphics[width=\linewidth]{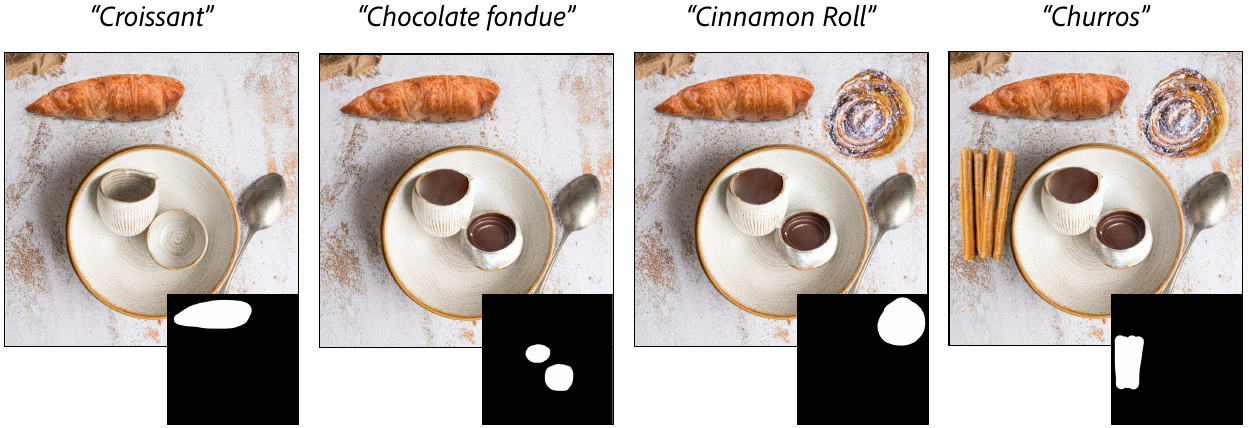} \\
             \includegraphics[width=\linewidth]{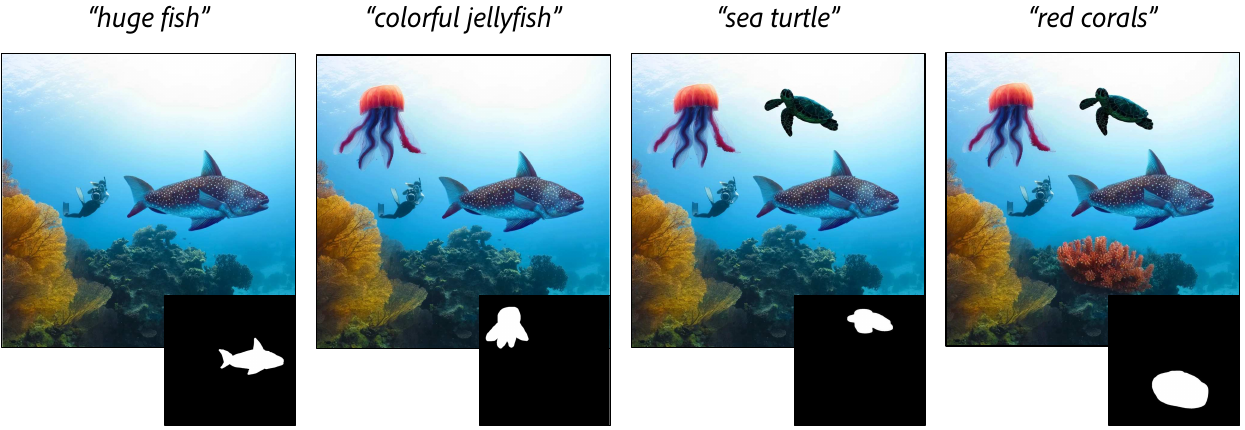} \\
             \midrule
             \includegraphics[width=\linewidth]{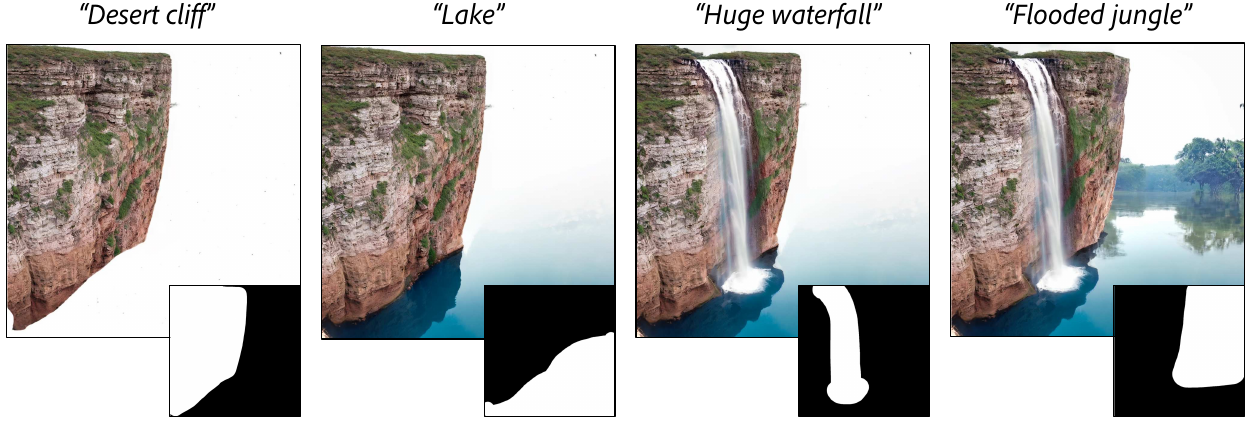} \\
             \includegraphics[width=\linewidth]{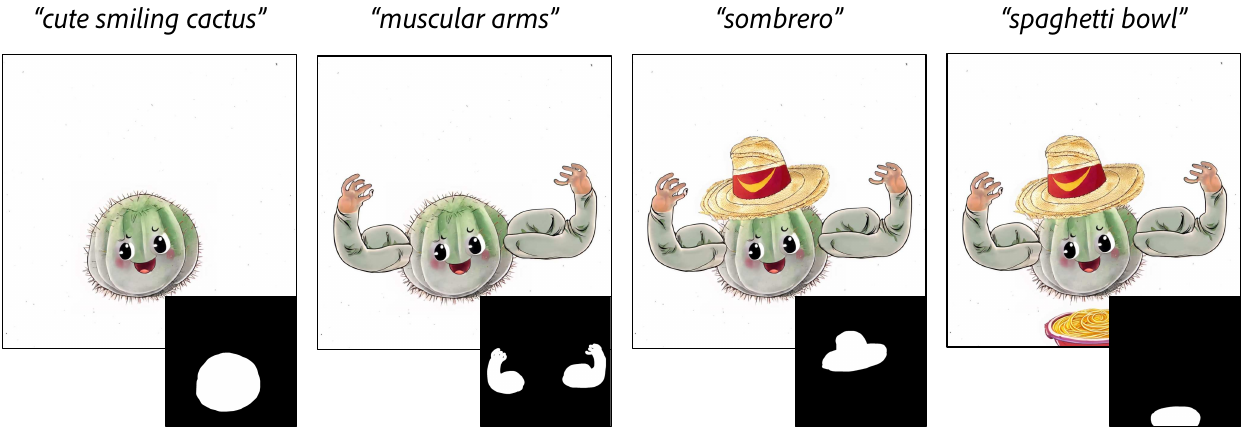} \\
        \end{tabular}
        \caption{
        Progressive image editing (top) and image generation (bottom) using \methodname.
        Each panel illustrates a generative progression compared to the preceding state of the canvas to its left.
        \methodname markedly accelerates local image edits (approximately $\times10$), rendering diffusion models more apt for user-in-the-loop applications.
        }
        \label{fig:gradual_generation}
        \vspace{-5mm}
    \end{figure}
\fi

%% file: figures/quant.tex
\begin{table}
    \centering
    \caption{
    Quantitative comparison of our method with the three baselines.
    We report zero-shot FID~\cite{heusel2017gans} and CLIPScore~\cite{hessel2021clipscore} on $10k$ images images from OpenImages~\cite{schuhmann2021laion}.
    Scores of SD2-crop~\cite{rombach2022high} and SDXL~\cite{podell2023sdxl} are not directly comparable and provided only for reference.
    }
    \begin{tabular}{lcc}
        Method & CLIP Score $(\uparrow)$ & FID $(\downarrow)$ \\
        \toprule
         \color{gray}{SD2-crop} & \color{gray}{0.21} & \color{gray}{6.95} \\
         \color{gray}{SDXL} & \color{gray}{0.21} & \color{gray}{6.88}  \\
         \midrule
         \baselinecrop & 0.19 & 9.35 \\
         \baselinefull & 0.19 & 7.38 \\
         \methodname (Ours) & 0.19 & 7.70
    \end{tabular}
    \label{tab:quant}
\end{table}

%% file: figures/qual_compare.tex
\begin{figure*}
    \centering
    \begin{tabular}{l}
         \includegraphics[width=0.9\linewidth]{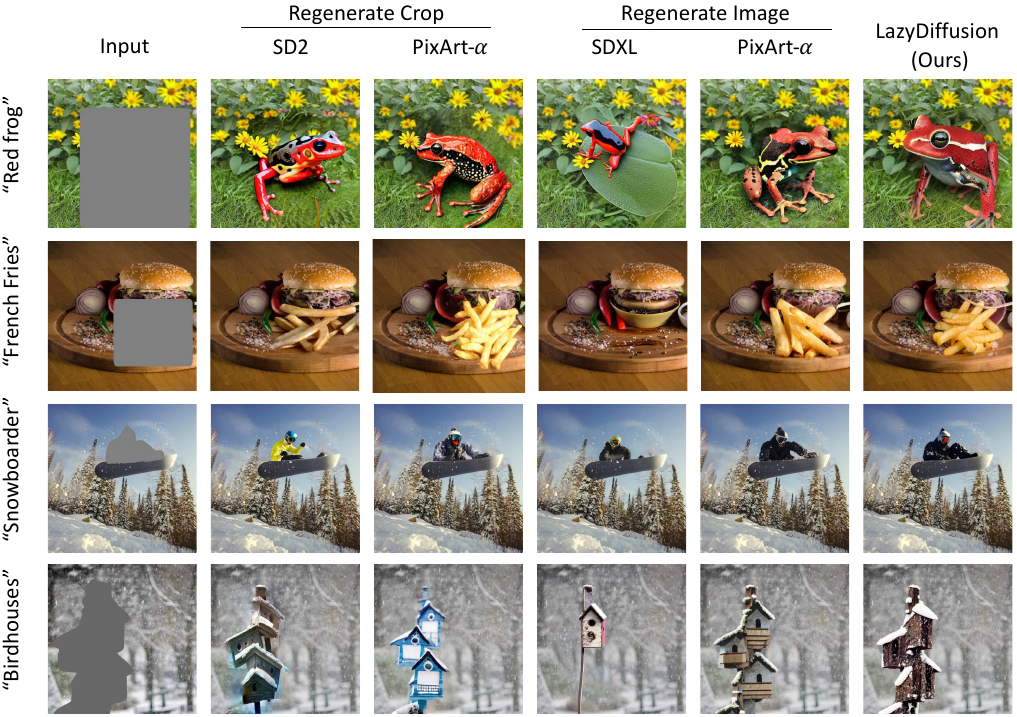} \\
         \midrule
         \includegraphics[width=0.9\linewidth]{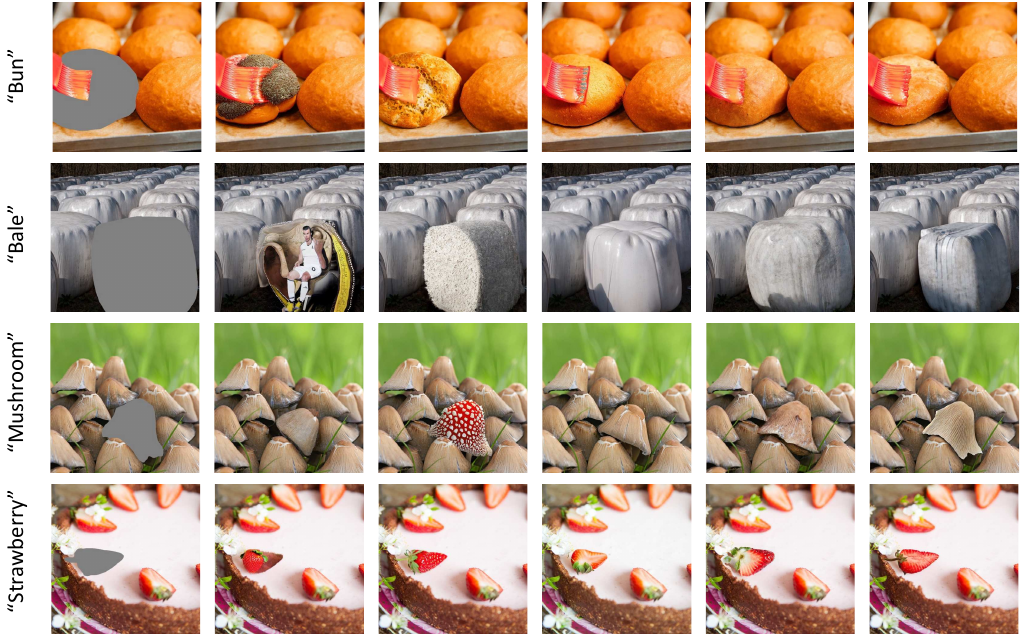}
     \end{tabular}
     \caption{
    Comparing Inpainting Results:
    (Top) Inpainting most objects requires relatively little semantic context. In such cases, all methods produce reasonably good results, even those processing only a tight crop.
    (Bottom) However, when inpainting an object closely related to others, such as one bun out of many, the inpainting model requires robust semantic understanding. Methods processing only a crop produce objects that may seem reasonable in isolation, but do not fit well within the greater context of the image.
    In contrast, \methodname adeptly leverages the compressed image context to generate high-fidelity results, comparable in quality to models regenerating the entire image and running up to ten times slower.
    \ifsingle \else Additional results are provided in~\cref{fig:supp_qual_high,fig:supp_qual_low,fig:random1,fig:random2}. \fi
     }
    \label{fig:qual_compare}
\end{figure*}

%% file: figures/SDEdit.tex
\begin{figure}
    \centering
    \includegraphics[width=\linewidth]{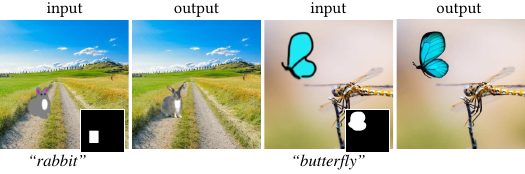}
    \caption{Our model readily supports additional forms of local conditioning. For example, similar to SDEdit~\cite{meng2021sdedit}, a user can draw a simplistic colored sketch, providing the model shape and color information.
    }
    \label{fig:sdedit}
\end{figure}

%% file: sections/6_conclusion.tex
\ifsingle
    \section{Societal impact}
    Generative models, including our work, can be used to produce misleading content that causes societal harm. Nevertheless, our work does not introduce unique concerns beyond this. We hope that increased public awareness and the development of more automatic tools for detecting generated media will mitigate these risks.
\fi

\section{Conclusions, limitations and future work}
\label{sec:conclusion}

We introduced a novel transformer-based encoder-decoder architecture for interactive image generation and editing using a diffusion model.
Our approach reduces the diffusion runtime by only generating the patches corresponding to the small region to synthesize, rather than the entire image.  
This is achieved through a global context encoder that summarizes the entire image once, outside the diffusion sampling loop, ensuring globally-consistent outputs.

Our method maintains the generation quality of state-of-the-art models, and reduces runtime costs proportionally to the size of the region to generate. 
This reduction in latency, particularly for small masks, transforms image generation into an interactive process by spreading the generation cost across multiple user interactions.

Our architecture does have some weaknesses. Despite operating outside the diffusion loop, the context encoder processes the entire image, posing a potential bottleneck for very high-resolution images due to its quadratic scaling in input size.
Addressing this limitation could enhance the scalability and applicability of our approach to larger and more intricate visual content.
We observed that occasionally, generated results have a subtle color shift compared to the visible image regions, leading to visible patch boundaries. 
While the Poisson blending post-processing methods discussed in Section~\ref{subsec:local-decoder} effectively mitigates these issues, future research is needed to identify a more principled and systematic solution.

\boldparagraph{Acknowledgement.}
We are grateful to Minguk Kang, Tianwei Yin and Wei-An Lin for technical suggestions, to Rotem Shalev-Arkushin for proofreading our draft and offering feedback, and to Yogev Nitzan for his help running the user study. This work was done while Yotam Nitzan was an intern at Adobe.

%% file: sections/X_appendix.tex
\vspace{-0.8mm}
\section{Supplementary Overview}
\ifsingle
    Supplementing our main paper, we include two additional files.
    First, we provide a \textbf{video} file comparing real-world interactions of users with \methodname and our baseline, \baselinefull.
    Second, we include this PDF document containing further experiments, results, and details.
\fi
In~\cref{sec:imagenet}, we conduct an ablation study, comparing our chosen architecture with possible alternatives.
Then, in~\cref{sec:supp_experiments}, we analyze our blending approach at post-processing and extend the qualitative evaluation from the main paper.
Finally, in~\cref{sec:details}, we offer additional implementation details, completing the paper.

\section{Architecture Design and Ablation}
\label{sec:imagenet}

Pivotal to our architectural design is compressing the visible context to fewer tokens and utilizing it within the diffusion decoder.
In the following section, we describe the experiments leading to our eventual design.

\subsection{Setting}
While text-based inpainting serves as the primary application demonstrated in this paper, \methodname is readily applicable to a range of other local generation applications. 
When designing our architecture in early stages of this work, we applied our method to unconditional inpainting
\cite{yu2019free,suvorov2021resolution} on ImageNet~\cite{deng2009imagenet} at $256\times256$ resolution, as this setting demands substantially less training time and resources.
We adopt the masking protocol from DeepFillV2~\cite{yu2019free}.
We use the same ViT XL/2~\cite{dosovitskiy2020image} backbone for our context encoder and adopt DiT XL/2~\cite{peebles2023scalable} for the diffusion transformer.
Note that the \pixart~\cite{chen2023pixartalpha} architecture, used in the main paper, is a straight-forward adaptation of DiT to support text conditioning.
Consequently, the architectures we describe next can seamlessly use both as backbones.

\subsection{Chosen design review}
\label{subsec:our_arch}
Recall that in our proposed architecture, discussed in Sec. 3, we selectively retain only encoder output tokens corresponding to the masked region, marked $\mathcal{T}_\text{hole}$.
This ensures that downstream decoder computation scales with the mask size rather than the image size. 
At time $t$, the decoder denoises tokens $\mathcal{X}_\text{hole}^{t}$ while conditioning on the retained context tokens.
We implement the conditioning by concatenating the context tokens to the noise tokens at the decoder's input.
Omitted from the main paper for clarity, we prepend a linear projection layer to the diffusion transformer backbone, projecting the concatenation of tokens to the decoder's hidden dimension $d$.
Other than the first layer, the diffusion transformer is then used \emph{as-is} to generate $k = | \mathcal{T}_\text{hole}| $ tokens.
Rewriting \cref{eq:diffusion} from the main paper with greater detail, a single denoising step reads as
\begin{equation}\label{eq:supp_diffusion}
    \mathcal{X}_\text{hole}^{t-1} =  \text{DiT}\left( \text{linear}( \mathcal{X}_\text{hole}^{t} \oplus \mathcal{T}_\text{hole});
    t, \mathbf{c}\right),
\end{equation}
where $\oplus$ denotes concatenation along the hidden dimension.
Transformers runtime scale quadratically with the number of tokens. 
Thus, the runtime of this architecture scales as $\mathcal{O}(k^2)$.
In this section, we refer to this architecture as the ``Concat Hidden'' variant.

\subsection{Alternative designs}
We next describe alternative designs with the goal of ablating the two core choices -- dropping visible tokens to compress context and conditioning through concatenation

\myparagraph{Full context designs}, utilizing the full set of $N$ encoder tokens $\mathcal{T}_\text{all}$ as context:
\vspace{2mm}
\begin{itemize}
    \item \baselinefull -- As described in the paper, we adapt DiT for inpainting using the GLIDE~\cite{nichol2021glide} conditioning approach.
    This model represents the common approach in local editing literature --  operates on the entire canvas thus seeing the full context but also re-generating the entire image.
    The runtime complexity of this variant scales as $\mathcal{O}(N^2)$.
    Note that $N >> k$.
    \vspace{0.5mm}
    \item \textit{Full-Context Cross-Attention} -- We add a cross-attention layer to the DiT block, between the self-attention and MLP layers.
    Other than the upstream activations, the cross-attention layer gets as input the \emph{full} encoder context tokens $\mathcal{T}_\text{all}$.
    Despite ``seeing'' the full context, the model generates only the $k$ masked patches.
    It's runtime scales as $\mathcal{O}(Nk)$.
\end{itemize}

\myparagraph{Compressed context designs.}
Comparable to our chosen design -- the following models utilize the masked tokens $\mathcal{T}_\text{hole}$ as context, generate only the masked region and have runtimes that scale with $\mathcal{O}(k^2)$. 
They differ in their mechanism to condition on the context tokens.
We experiment with simple conditioning approaches that are applied near the input level.
This prevents designs from being tightly coupled with the specific backbone architecture, which we anticipate would facilitate easier adaptation to future diffusion transformers.
\vspace{2mm}
\begin{itemize}
    \item \textit{Concat Length} -- The sets of tokens are concatenated over the sequence length, rather than hidden dimension.
    This requires the two sets of tokens to have the same hidden dimension.
    To this end, we first linearly project the context tokens to the decoder's hidden dimension $d$.
    Formally, a single denoising step is done by
    \begin{equation}\label{eq:cat_length}
        \mathcal{X}_\text{hole}^{t-1} =  \text{DiT}\left( [ \mathcal{X}_\text{hole}^{t}, \text{linear}(\mathcal{T}_\text{hole})];
        t, \mathbf{c}\right),
    \end{equation}
    where $[\cdot,\cdot]$ represents the sequence-length concatenation.

    \item \textit{Weighted Sum} --  An additional weight $w \in \mathbb{R}^d$ is learned, and the input to DiT is a weighted sum of the two sets of tokens, formally 
    \begin{equation}\label{eq:weighted_sum}
        \mathcal{X}_\text{hole}^{t-1} =  \text{DiT}\left( \mathcal{X}_\text{hole}^{t} + w * \text{linear}(\mathcal{T}_\text{hole});
        t, \mathbf{c}\right).
    \end{equation}

    \item Compressed-Context Cross-Attention -- We again add a cross-attention layer, but here it attends only to the reduced set of tokens $\mathcal{T}_\text{hole}$.
    To better resemble other designs in this category, incorporating the conditioning near the input, we add the cross-attention layer only to the first DiT block.
\end{itemize}

\subsection{Configurations}
DiT's FLOPs are strongly negatively correlated with FID, across different configurations~\cite{peebles2023scalable}.
To facilitate direct comparison, we slightly adjust the XL/2 configuration for the $\mathcal{O}(k^2)$ variants so that their FLOP counts are similar.
We provide the exact hyperparameters used with each variant in~\cref{tab:arch_config} and the resulting FLOP counts as a function of mask size are in~\cref{fig:arch_flops}.
As can be seen, \textit{Concat Hidden}, \textit{Weighted Sum} and the \textit{Compressed-Context Cross-Attention} have comparable FLOPs on the entire spectrum ranging from mask ratio of 10\% to 100\%.
For full masks, the \textit{Concat Hidden}, \textit{Weighted Sum} variants use $0.4\%$ and $0.6\%$ more FLOPs than \baselinefull, respectively.
This implies that our conditioning introduces negligible overhead and is well suited for using larger masks with no apparent downside.
The other three variants have strictly greater FLOP counts.
\input{figures/supp_archiecture_flops}

\subsection{Results}
We track the FID~\cite{heusel2017gans} scores across 500K training iterations for all decoder designs and present the results in \cref{fig:arch_ablation}.

Initially, we observe that ``Concat Hidden'' and ``Weighted Sum'' notably outperform all other variants.
We attribute this superior performance to the explicit one-to-one context provided by these approaches.
In both cases, each noise token is directly conditioned on the corresponding context token.
In contrast, other methods require the decoder to extract context from a set of encoder tokens, which appears to be more challenging despite the use of positional embedding and more expressive mechanisms such as cross-attention.

Furthermore, we note that the more computationally intensive baselines, which leverage additional context, do not yield better results. Specifically, in the two cross-attention variants, the one that uses compressed context is superior to the one using full context.
Our attempts to improve the performance of the \baselinefull baseline by using a context encoder and a ``Concat Hidden'' based conditioning were futile; only dropping the visible context tokens was effective.
We speculate that incorporating the full context imposes additional complexity on the decoder's task.
In comparison, with \methodname, the information bottleneck encourages the context to be expressive but selective, allowing the decoder to ``concentrate'' on synthesis only.

Interestingly, in the text-conditioned setting, \methodname is not superior in terms of quality to \baselinefull.
This disparity might be explained by the lower level context required for unconditional inpainting, which primarily involves continuing surrounding textures, compared to the semantic context required for generating novel objects.

\subsection{Implementation details}
We train and sample all models with the EDM~\cite{karras2022elucidating} diffusion formulation.
We use Stable Diffusion's~\cite{rombach2022high} public latent VAE.
We train the encoder and decoder jointly from scratch, on 8 NVIDIA A100 GPUs, using global batch size of 256, using the AdamW~\cite{loshchilov2017decoupled} optimizer with constant learning rate of $10 ^ {-4}$.
We sample using 40 denoising steps and classifier-free guidance scale of $4.0$.
Other details are the same as in the text-conditioned setting and are detailed in the main paper or in~\cref{sec:details}.

\section{Additional Experiments and Results}
\label{sec:supp_experiments}

\subsection{Blending}
\methodname generates only the masked regions of the latent image.
To achieve the final desired results, these regions must be composited with the visible image regions and decoded into an image.
Initially, we naively blend the generated latent with the latent of the input image, as described in \cref{eq:output-latent} in the main paper.
However, we observe that passing the blended latent through the latent decoder 
$\mathcal{D}$ occasionally results in poorly harmonized images, characterized by faintly visible seams between the generated and visible regions.
This phenomenon was previously noted by Zhu et al. \cite{zhu2023designing} when performing local editing with Stable Diffusion \cite{rombach2022high}.
It is conjectured that the latent encoding loses subtle color information, hindering image harmonization.
In response, Zhu et al. proposed an alternative latent decoder that additionally conditions on the masked input image $I \odot (1-M)$ itself and is also significantly larger.
Specifically, their decoder runs for $800$ms, $4.5\times$ longer than the ``vanilla'' Stable Diffusion latent decoder.

In our experiments, we find that simply performing Poisson blending~\cite{perez2003poisson} in pixel space achieves comparable results, while running only for $35$ms on average. 
Therefore, we introduce a Poisson blending post-processing step to our pipeline. 
We demonstrate the harmonization issue and compare the two approaches in \cref{fig:supp_blend}.
\input{figures/supp_blend_ablation}

\subsection{Additional Results}
In \cref{fig:supp_qual_high,fig:supp_qual_low}, we extend \cref{fig:qual_compare} of the main paper and provide more qualitative samples comparing \methodname with the four baselines -- \baselinecrop, SD2-crop, \baselinefull and SDXL.
We find that \methodname is mostly comparable to \baselinefull and SDXL even when inpainting objects that require high semantic context, despite using a compressed context and running up to $10\times$ faster.

Finally, in~\cref{fig:random1,fig:random2} we provide a non-curated set of results, with masks and text prompts produced automatically by the segmentation and captioning models.
The main challenge we observe from these results is that the model partially ignores the text when it conflicts with the shape of the mask.
For example, the hamburger in~\cref{fig:random1} is generated without a hat.

\input{figures/supp_low_context}
\input{figures/supp_high_context}
\input{figures/supp_random}

\section{Additional Details}
\label{sec:details}

\myparagraph{Evaluation.}
We compute FID~\cite{heusel2017gans} using clean-fid~\cite{parmar2021buggy}.
For CLIPScore~\cite{hessel2021clipscore}, we report the ``local'' version that takes as input a crop around the generated object and the local text, describing the object. 
This approach was previously advocated by Wang \etal~\cite{wang2023imagen} and is more suitable for image inpainting than using the full image and text caption for the entire image.

\myparagraph{Architecture.}
As described in the main paper, we initialize our decoder with \pixart's publicly released weights.
Our decoder has an additional linear layer, introduced in~\cref{subsec:our_arch}, that projects the concatenation of context and noise tokens to the decoder's hidden dimension $d$.
We initialize this layer such that it outputs the noise tokens in its input and ignores the context.
This ensures that at initialization, if given a full mask and thus operates on all tokens, our results are exactly equivalent to \pixart's.

\myparagraph{Data.}
As discussed in the paper, we adopt a data processing pipeline similar to that of SmartBrush~\cite{xie2023smartbrush}.
Specifically, our masks are originally produced by an entity segmentation model~\cite{qi2022open} and are dilated to simulate the rough and inaccurate masks created by users.
First, with probability of 20\% we replace the segmentation mask with a rectangular mask corresponding to a bounding box.
Regardless, we dilate the mask by first performing Gaussian Blurring and thresholding the output.
The size of the Gaussian kernel is sampled uniformly from $[\text{image size} / 15, \text{image size} / 5]$ and its standard deviation along X and Y is sampled uniformly and independently from $[3,17]$. The threshold is sampled uniformly from $\{10^{-1}, 10^{-2}, 10^{-3}, 10^{-4}\}$.

%% file: figures/supp_archiecture_flops.tex
\begin{table}[]
    \ifsingle
        \setlength{\tabcolsep}{12pt}
    \fi
    \centering
    \caption{
    Hyperparameters configuration for all architecture designs.
    Starting from DiT's XL/2 configuration, we slightly adapt the hyperparameters to ensure FLOP counts of $\mathcal{O}(k^2)$ are comparable.
    }
    \label{tab:arch_config}
    \begin{tabular}{lccc}
         \makecell{Runtime \\ Complexity} & Model & Layers & \makecell{Hidden \\ Dimension} \\
         \toprule
         \vspace{-4mm}
         \multirow{4}{*}{$\mathcal{O}(k^2)$} &
                     &
                     &
                     \\
         & Concat Hidden & 28 & 1152 \\
         & Weighted Sum & 28 & 1152 \\
         & Concat Length & 24 & 1024 \\
         & Cross Attention & 26 & 1152 \\
         \midrule
         $\mathcal{O}(Nk)$ & Cross Attention & 28 & 1152 \\ 
         $\mathcal{O}(N^2)$ & \baselinefull & 28 & 1152 \\
         \bottomrule     
    \end{tabular}
\end{table}

\begin{figure}
    \centering
    \begin{subfigure}{\linewidth}
        \centering
        \includegraphics[width=\linewidth]{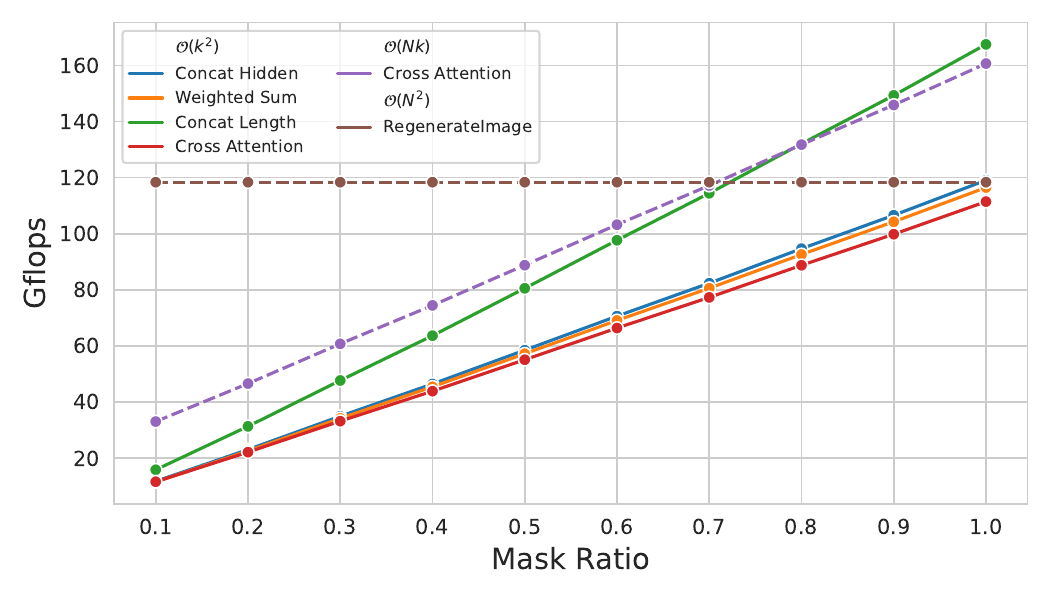}
        \caption{}
        \label{fig:arch_flops}
    \end{subfigure}
    \begin{subfigure}{\linewidth}
        \centering
        \includegraphics[width=\linewidth]{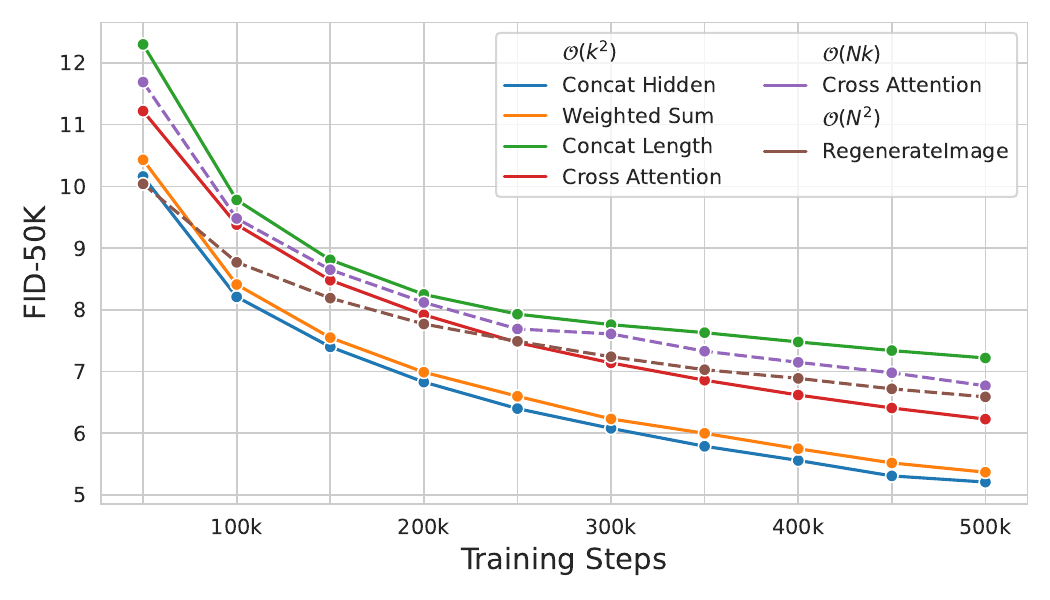}
        \caption{}
        \label{fig:arch_ablation}
    \end{subfigure}
    \caption{
    Comparing the various architecture designs in terms of (a) FLOPs and (b) quality, measured via FID~\cite{heusel2017gans}.
    Solid lines represent variants of our approach -- the encoder outputs a compressed context and the decoder generates only the masked region.
    Dashed lines represent mechanisms in which the decoder is conditioned on the full image context and either generates the masked region or the entire image. 
    The latter is the approach taken by existing inpainting approaches~\cite{rombach2022high}.
    The runtime complexities of different approaches is noted in the legend.
    As can be seen, conditioning each generated token directly on its corresponding compressed context token, as done for the ``Concat Hidden'' and ``Weighted Sum'' variants, leads to superior performance, despite using fewer FLOPs than competing approaches.
    }
\end{figure}

%% file: figures/supp_blend_ablation.tex
\begin{figure}
    \centering
    \includegraphics[width=0.98\linewidth]{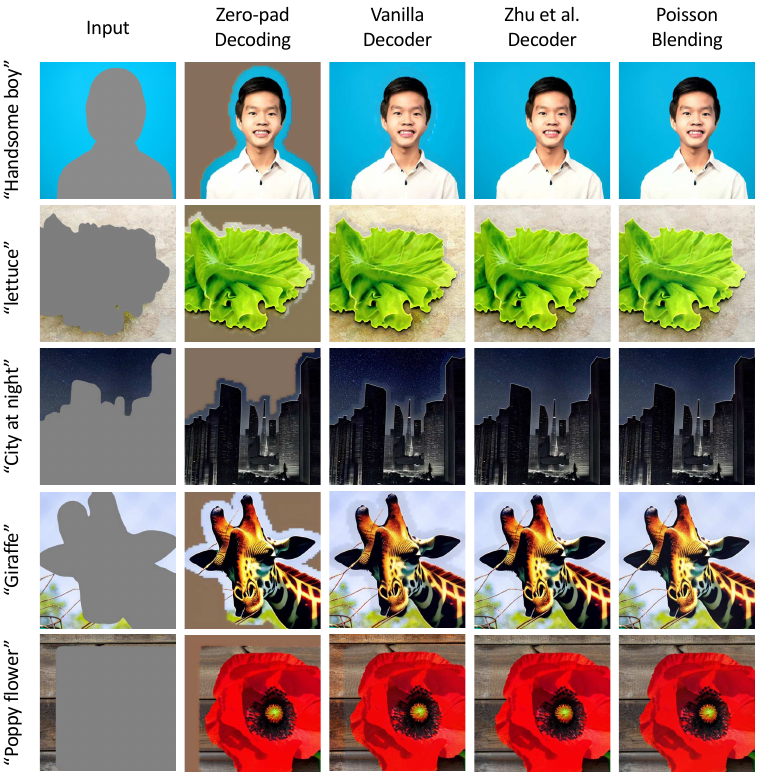} \\
     \caption{
     From partial latent generation to inpainted image.
     The ``zero-pad decoding'' column is produced by decoding the incremental generation with zero padding, demonstrating the object in isolation.
     To produce the desired composited image, we blend the incremental generation with the latent input.
     This occasionally leads to visible seams and lack of color harmonization as seen in the ``vanilla decoder'' column.
     This issue can be solved using the latent decoder proposed by Zhu \etal~\cite{zhu2023asymmetric} or with Poisson blending~\cite{perez2003poisson}.
     We recommend zooming in to better view the seams or lack thereof.
     }
    \label{fig:supp_blend}
\end{figure}

%% file: figures/supp_low_context.tex
\begin{figure*}
    \centering
    \includegraphics[width=0.98\linewidth]{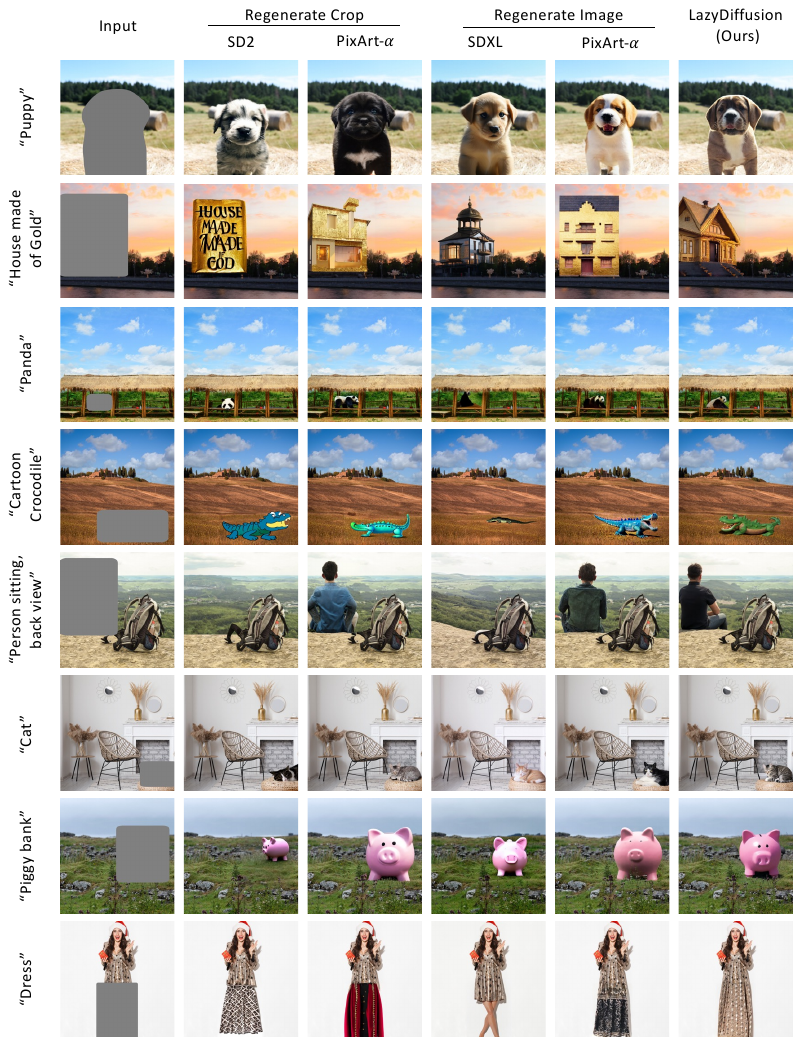} \\
     \caption{
     Comparing inpainting results on objects that require modest context, similar to Fig. 7(Top).
     All models usually produce reasonably good results.
     Occasionally, SDXL~\cite{podell2023sdxl} and SD2~\cite{rombach2022high} do not generate anything -- a result of their usage of random masks rather than object-level masks~\cite{xie2023smartbrush,wang2023imagen}.
     }
    \label{fig:supp_qual_low}
\end{figure*}

%% file: figures/supp_high_context.tex
\begin{figure*}
    \centering
    \includegraphics[width=0.98\linewidth]{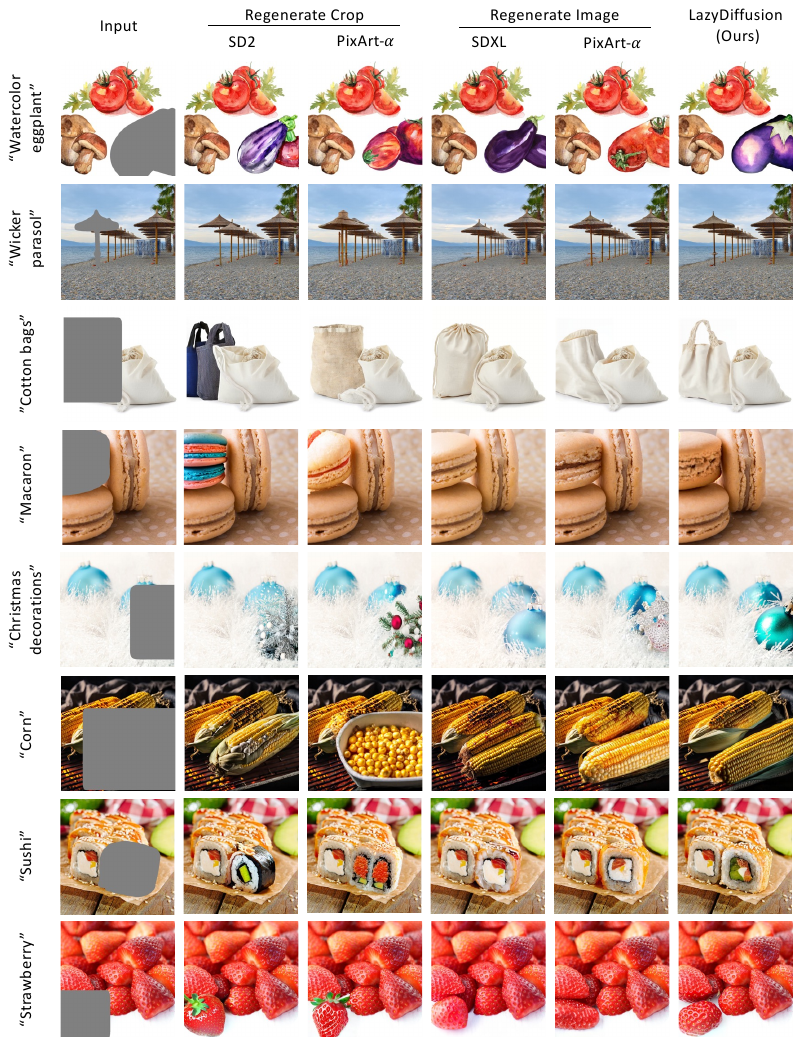} \\
     \caption{
     Comparing inpainting results on objects that have close semantic relationship with the observed canvas, similar to Fig. 7(Bottom).
     Approaches that only process a crop may generate objects that appear reasonable on their own but lack coherence within the broader context of the image.
     In contrast, \methodname produces results comparable to those produces by methods regenerating the entire image.
     Occasionally, \methodname does not fully utilize the visible context. 
     For instance, our ``sushi'' result accurately depicts the orange wrap and sesame seeds on top, consistent with other sushi in the roll, but it features a different filling.
     }
    \label{fig:supp_qual_high}
\end{figure*}

%% file: figures/supp_random.tex
\begin{figure*}
    \centering
    \includegraphics[height=0.99\textheight]{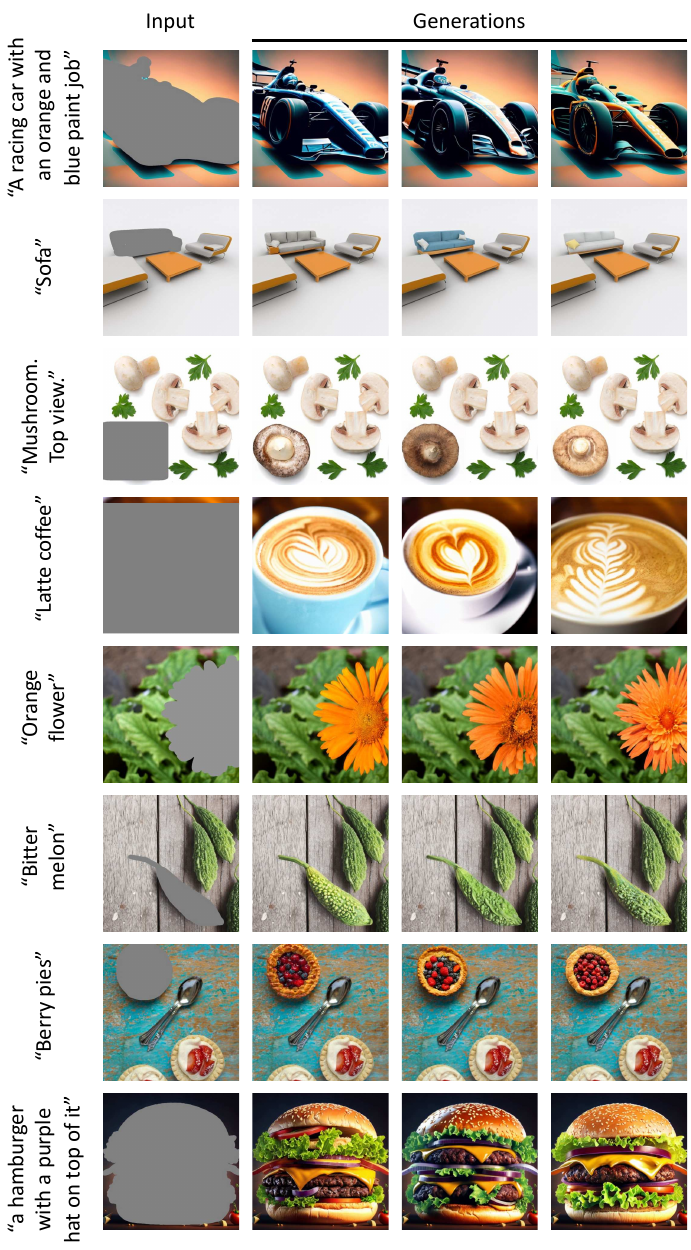} 
    \caption{
    A random set of results produced by \methodname.
    For each input we produce three outputs from different random seeds.
    }
    \label{fig:random1}
\end{figure*}

\begin{figure*}
    \centering
    \includegraphics[height=0.99\textheight]{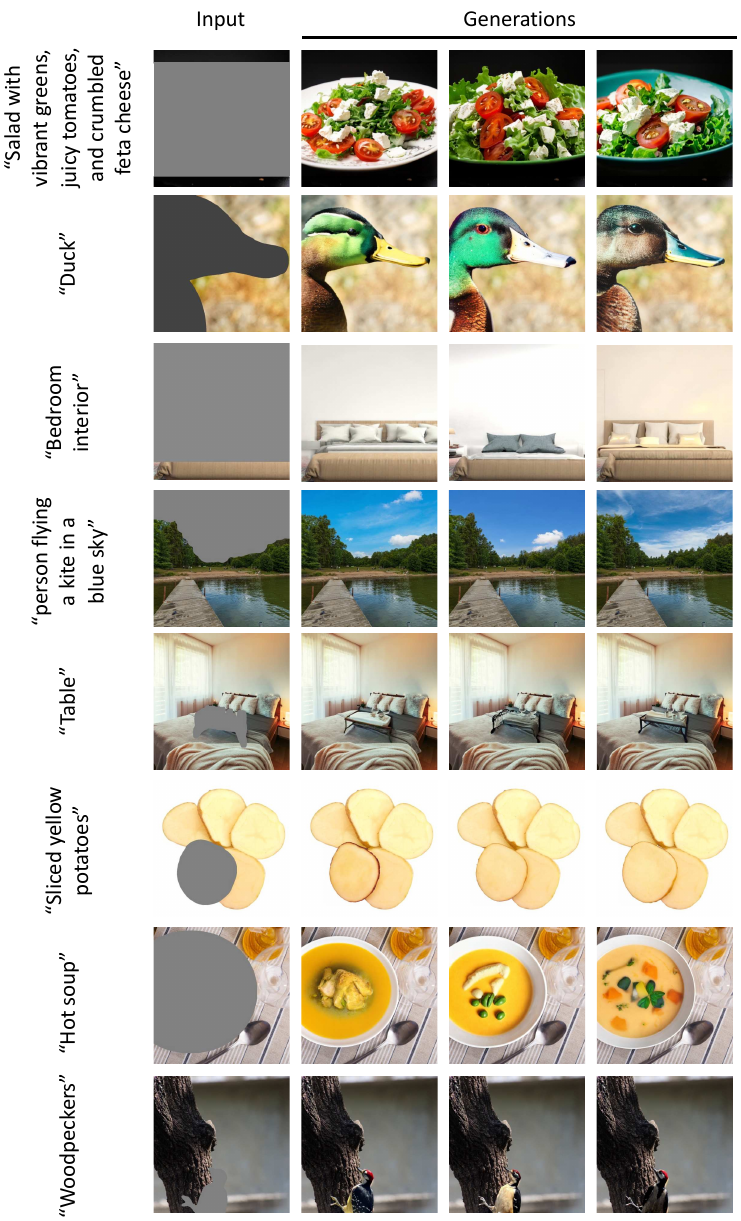} 
    \caption{
    A random set of results produced by \methodname.
    For each input we produce three outputs from different random seeds.
    }
    \label{fig:random2}
\end{figure*}